# Detecting Unexpected Obstacles for Self-Driving Cars: Fusing Deep Learning and Geometric Modeling

Sebastian Ramos[1,2], Stefan Gehrig[1], Peter Pinggera[1], Uwe Franke[1] and Carsten Rother[2]

*Abstract*— The detection of small road hazards, such as lost cargo, is a vital capability for self-driving cars. We tackle this challenging and rarely addressed problem with a vision system that leverages appearance, contextual as well as geometric cues.

To utilize the appearance and contextual cues, we propose a new deep learning-based obstacle detection framework. Here a variant of a fully convolutional network is used to predict a pixel-wise semantic labeling of *(i) free-space*, *(ii) on-road unexpected obstacles*, and *(iii) background*. The geometric cues are exploited using a state-of-the-art detection approach that predicts obstacles from stereo input images via model-based statistical hypothesis tests.

We present a principled Bayesian framework to fuse the semantic and stereo-based detection results. The mid-level Stixel representation is used to describe obstacles in a flexible, compact and robust manner.

We evaluate our new obstacle detection system on the *Lost and Found* dataset, which includes very challenging scenes with obstacles of only 5 cm height. Overall, we report a major improvement over the state-of-the-art, with relative performance gains of up to 50%. In particular, we achieve a detection rate of over 90% for distances of up to 50 m. Our system operates at 22 Hz on our self-driving platform.

## I. INTRODUCTION

The detection of small-sized and unexpected road hazards, such as lost cargo, is a very demanding and important task for human drivers. US traffic reports show that approximately 150 people are killed annually due to lost cargo on the road [2]. Consequently, self-driving cars operating on public roads have to be able to avoid running over such obstacles by all means. Considering the resulting requirements on environment perception systems, this task represents a challenging and so far largely unexplored research topic. In order to close the gap between semi-autonomous and fully autonomous driving capabilities, a robust and accurate solution to this problem has to be found.

Besides rather expensive high-end time-of-flight sensors, RGB cameras represent a promising means to approach this task due to their high resolution. Previous systems that addressed this problem make use of stereo cameras in order to take advantage of geometric cues for the detection and 3D localization of small generic obstacles [1].

However, safe operation of fully autonomous vehicles requires an extremely high degree of accuracy and reliability with regard to environment perception systems. In the particular case of detecting objects that cover only small

[1] S. Ramos, S. Gehrig, P. Pinggera and U. Franke are with the Environment Perception Dept., Daimler AG R&D, Sindelfingen, Germany {firstname.lastname}@daimler.com

[2] C. Rother is with the Computer Vision Lab Dresden, TU Dresden, Dresden, Germany carsten.rother@tu-dresden.de

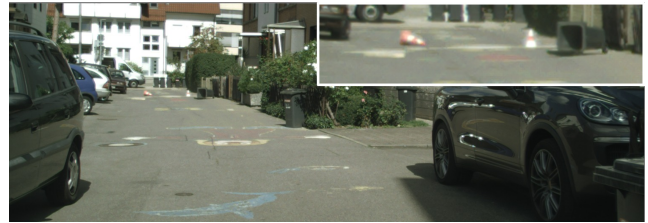

Input image and close-up showing four small but critical obstacles.

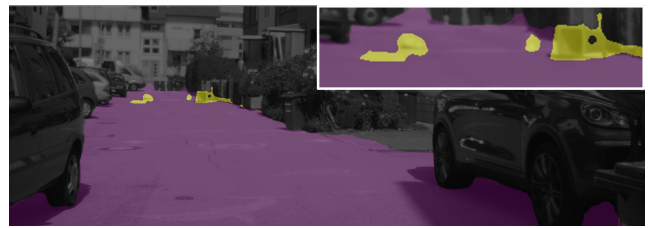

Pixel-wise semantic labeling result of our appearance-based detector.

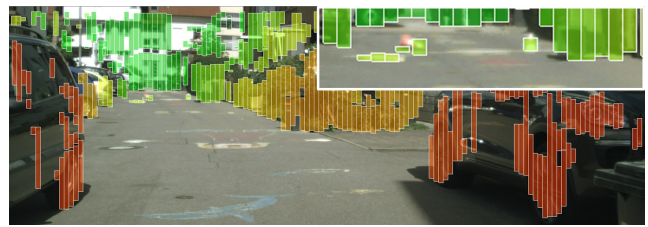

Stixel representation of the detection results of the probabilistic fusion system (semantics + geometry).

Fig. 1. Exemplary scene from the *Lost and Found* dataset [1] and corresponding results of the proposed system. All relevant obstacles are detected successfully despite their challenging size and appearance. Semantic labels are color-coded by class, final detections are color-coded by distance.

image regions and occur in all possible shapes and dimensions, utilizing geometric cues alone is not sufficient. Here machine learning techniques that leverage appearance cues represent a promising complementary approach to the current geometry-only systems. However, due to the high variability in appearance of both objects and road surfaces, traditional supervised learning systems are bound to encounter problems in practice.

Recently, research on artificial intelligence has experienced a substantial boost due to advances in deep learning methods [3]. In the field of computer vision, deep neural networks and especially deep Convolutional Neural Networks (CNNs) have emerged as an extremely powerful tool to learn feature representations for a large variety of tasks from low-level (e. g. image segmentation) to high-level vision (e. g. object detection and classification). Arguably, an even greater strength of CNNs lies in their ability to exploit contextual information learned from large quantities of training data.

In our application, small obstacles lying on the road share one common property, which is context, that can be exploited despite the inherent variance in shape, size and appearance. In particular, such obstacles are generally surrounded by drivable road area and thus stand out as an anomaly from an otherwise regular or piece-wise homogeneous surface.

In this paper, we introduce a new deep learning-based approach to detect generic, small, and unexpected obstacles on the road ahead. In this way we demonstrate how the ability of CNNs to learn context and generalize information from training data can overcome one of their main open problems: handling outliers and the "open world". This refers to the tasks of correctly handling novel, previously unseen objects as well as modeling a suitable background class.

Furthermore, we present a probabilistic fusion approach that combines our learning-based detection method with the currently best performing stereo-based system for this task. In our in-depth evaluation we report a relative increase in detection rate by 50% compared to the state-of-the-art, with a simultaneous decrease of 13% in false positives.

The remainder of this paper is organized as follows: Section II reviews related work on the considered topic, covering geometric model-based as well as machine learning methods. In Section III the proposed detection approaches are described in detail. Section IV offers an in-depth evaluation, focusing on a quantitative analysis of detection performance and false positive rates, followed by qualitative results as well as discussions. The final section provides a summary and a brief outlook.

## II. Related Work

Obstacle detection has been a heavily researched topic for decades, spanning a variety of application areas. However, little work has focused on the detection of small, generic and unexpected obstacles in driving scenarios. We mainly consider camera-based methods for the detection and localization of such generic obstacles in 3D space, using stereo camera setups with small baselines (i. e. less than 25 cm).

Classical geometry-based obstacle detection schemes are mostly based on the flat-world-assumption, modeling free-space or ground as a single planar surface and characterizing obstacles by their height-over-ground (e.g. [4]). Geometric deviations from this reference plane are often estimated either from image data directly [5] or via mode extraction from the v-disparity histogram on multiple scales [6]. In order to cope with deviations from the flat-world-assumption, more sophisticated ground profile models have been introduced, e.g. piece-wise planar longitudinal profiles [7] or splines [8].

A recent survey [9] presents an overview of several stereo-based generic obstacle detection approaches that have proven to perform well in practice. The methods are grouped into different obstacle representation categories including Stixels [10], Digital Elevation Maps (DEM) [11] and geometric point clusters [12]. DEMs have been recently used in the agricultural field for obstacle detection, exploiting a set of rules per grid cell to determine obstacle candidates [13]. We select the Stixel method [10] to serve as one of our baselines during our experimental evaluation. The Stixel algorithm discriminates between an estimated global ground surface model and a set of rectangular vertical obstacle segments, providing an efficient and robust representation of the 3D scene.

The methods presented above all rely on precomputed stereo disparity maps and obtain reliable detection results based on generic geometric criteria. They perform well for detecting medium-sized objects at close to medium range, with reduced performance for longer ranges and smaller obstacles.

Our present work utilizes the obstacle detection approach of [1] as one of the main system components. This method was originally presented in [14] within the scope of high-sensitivity long range detection tasks. It was extended in [1] and successfully applied to the detection of small and challenging road hazards. The stereo-based detection system performs independent, local tests, comparing obstacle against free-space hypotheses using constrained plane models, which are optimized directly on the underlying image data. The method however does require a certain minimum obstacle height in the image (i. e. in pixels), limiting the detection range for very small obstacles with the given camera setup to less than 40 m. Additionally, the *Lost and Found* dataset was introduced in this work.

The appearance-based detection of unexpected obstacles has not been widely investigated. To the best of our knowledge, the recent work of [15] first explored this particular problem from a machine learning perpective. This work focuses on the detection of regions with anomalous appearance on the road using a Restricted Boltzmann Machine neural network. This network is trained to reconstruct the appearance of the road in a given scenario. By subtraction of the observed and the reconstructed road appearances, the proposed method is able to segment small generic objects on the road. As mentioned by the authors, this work is at an early stage and has limitations to generalize to roads with appearances differing from the training examples, a main drawback for the considered application scenario.

Although not directly focused on the detection of unexpected road hazards, several works have been exploring machine/deep learning techniques for the segmentation of the free (drivable) space from visual data in the context of autonomous vehicle navigation. [16] presents an approach that combines features extracted from a deep hierarchical network and a real-time online classifier to predict traversability in off-road environments. This work uses a near-range stereo vision system to automatically collect ground truth data, in order to adapt the online classifier to changes of the free-space in front of a moving robot.

Following a very similar idea, [17] recently proposed a method for free-space detection in urban environments with a self-supervised and online trained Fully Convolutional Network (FCN) [18]. In this case, the automatic ground truth collection process is based on a combination of disparity maps and Stixel-based ground masks computed from a calibrated stereo-pair.

A different piece of related work from the deep learning community has recently been presented in [19]. This paper investigates the problem of adapting deep neural networks for open set recognition. Its goal is to provide deep learning-based recognition system with the ability to reject unseen/unexpected classes at test time. Further research in this direction is required for a direct application in the type of tasks considered in the present paper.

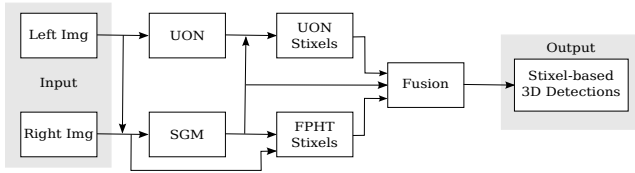

Fig. 2. Overview of the proposed processing pipeline. Appearance and context are exploited by our *Unexpected Obstacle Network (UON)*, while geometric cues are extracted in the stereo-based pipeline consisting of *Semi-Global-Matching (SGM)* and the *Fast Direct Planar Hypothesis Testing (FPHT)* obstacle detector. The mid-level Stixel representation is used to represent detected objects.

Last but not least, [20] recently presented the Semantic Stixels scene model. This model proposes a joint inference of geometric and semantic visual cues to generate a compact model representation of urban scenes for the perception requirements of self-driving cars. The geometric cues are included as disparity maps computed from a stereo vision system, while the semantic cues come as pixel-wise scene labels estimated from a deep learning approach. This model provides a compact yet rich scene representation and works very well at the detection of the most common participant of urban scenarios. However, one of the main limitations of this approach lies in the detection of small objects of an unknown, i.e. previously unseen, class. This method is included as a baseline in Section IV.

## III. METHODS

The processing pipeline of our complete detection system is illustrated in Fig. 2. An input image pair is processed using two parallel and independent channels to initially perform the pixel-level semantic reasoning as well as depth estimation. Subsequently, two independent 3D Stixel-based obstacle representations are obtained, the first exploiting the semantic and the second the geometric properties of the scene. Finally, both channels are fused in a probabilistic manner to generate an overall 3D Stixel representation of the potential obstacles on the road.

### A. Appearance-Based Semantic Detection: Unexpected Obstacle Network (UON)

Our first detection method relies on a pixel-wise semantic labeling of a single input image. This labeling leverages visual appearance and context via a deep learning approach, assigning every pixel of the image to a certain class.

For our particular problem, the complete drivable free-space in front of the vehicle has to be identified. At the same time, obstacles of any kind located inside this relevant region have to be detected. Therefore, we define the following set of classes that our learning system has to be able to distinguish: *free-space*, *on-road unexpected obstacles* and *background*.

The definition of this particular set of classes allows us to meet the demands of our application by exploiting the power of deep learning methods, i.e. the learning of context. Learning that relevant obstacles have a common contextual property, being of small dimensions and surrounded at least partly by free-space, gives our classifier the ability to detect small objects, at large distances and of diverse appearance. It turns out that it is able to generalize far beyond its training data, a critical fact in view of the innumerable possible corner cases that our system might have to face.

The *background* class is defined as any image region of no relevance to our application (e.g. sky regions, buildings, etc). Note that this class also contains all standard obstacles in urban traffic scenes (e.g. cars, pedestrians, etc). Due to their size and common occurrence, such objects can safely be assumed to be handled by general purpose perception algorithms such as [20].

In order to gain the required semantic understanding we make use of a Fully Convolutional Network (FCN) [18] with a GoogLeNet network architecture [21]. This combination provides a high classification accuracy at a relative low computational cost and GPU memory demands, a suitable trade-off for the hardware limitations of self-driving cars. We refer to our network as *Unexpected Obstacle Network (UON)*.

Following [18] we replace the last fully connected layer of the GoogLeNet architecture and include skip layers and a deconvolution layer with a bilinear weight filter to match the desired output size. The FCN outputs a probability map per class for all pixels of the input image. These maps, also referred to as heat maps, are calculated by a softmax normalization layer and are then further processed by a last argmax layer.

For training this network we combine two recently presented datasets for semantic labeling, namely the *Lost and Found* [1] and the *Cityscapes* [22] datasets. The combination of these complementary datasets provides examples of challenging suburban street scenes featuring small on-road obstacles as well as examples of complex inner-city scenes with many traffic participants and a variety of road markings, which helps to improve the robustness of our system.

In order to compensate for the imbalance of pixels of each class in the described dataset combination, we include a weighting factor for the *on-road unexpected obstacles* class within the cross-entropy loss of the softmax classifier. This proves to be crucial for achieving the desired trade-off between true detections and false alarms.

**UON-Stixel Generation:** The UON-Stixels are generated from the argmax image of the FCN output. In a first step, the network output is aligned with the Stixel width, computing the median of each class within one row of a Stixel. Thereby we horizontally downsample the argmax image by a factor equal to the defined Stixel width. A Stixel is generated if an obstacle-label occurs in the argmax image. The Stixel is expanded vertically until the labeling changes within the column. In this way even Stixels on small obstacles at large distances (e.g. of just a couple of pixels height in the image) are considered. In order to assign a 3D location to each generated Stixel, we average the underlying disparities from the disparity map output of our real-time implementation of Semi-Global Matching (SGM) [23].

*B. Stereo-Based Geometric Detection: Fast Direct Planar Hypothesis Testing (FPHT)*

As the second main component in our detection system, we apply the stereo-based *Fast Direct Planar Hypothesis Testing (FPHT)* approach of [1]. It was shown to perform exceedingly well at the challenging task of detecting small generic obstacles in complex urban road scenes.

This detection algorithm performs statistical hypothesis tests on small local patches distributed across the stereo images, where the test statistic is based directly on normalized input image residuals. Free-space is represented by the null hypothesis, while obstacles correspond to the alternative hypothesis. The hypotheses are characterized by constraints on the orientations of local 3D plane models, implicitly handling non-flat ground surfaces. For each local patch a generalized likelihood ratio test is formulated and solved by optimizing the hypothesis model parameters in disparity space. By determining a suitable decision threshold, obstacle points are thus reliably detected and localized in 3D space.

**FPHT-Stixel Generation:** The detected obstacle points are clustered and reshaped into Stixel-like groups, yielding a compact and flexible mid-level obstacle representation. For further details we refer to [1], [14].

Note that in the present work we make use of the algorithm variant denoted as "FPHT-CStix (downsampled)" in [1], and interchangeably refer to it as FPHT or FPHT-Stixels.

*C. Fusing Appearance- and Stereo-Based Detection*

In this section we describe our probabilistic fusion approach to combine UON-Stixels and FPHT-Stixels, also taking into account raw stereo disparity information.

**AND/OR Fusion Baselines:** As baselines to estimate optimal false positive and detection rates, we consider two simple fusion schemes represented by logical AND (optimal false positive rate) and OR (optimal detection rate) operators. The operators are applied to the UON and FPHT outputs, yielding a resulting Stixel if either one (OR) or both (AND) systems report a Stixel at a certain position. Corresponding Stixel are determined by an overlap of at least 50%. If corresponding Stixels are found, only the ones provided by FPHT are kept. A more elaborate refinement of the Stixel dimensions is conceivable, but since the exact position is less important than the actual existence we skip this step.

**Probabilistic Fusion:** We estimate the existence of a Stixel within a Bayesian framework. For simplicity, in the following we refer to the probability of existence for a Stixel as *confidence*. Computation starts with the OR-fused list of Stixels, where for each Stixel three cues are computed: FPHT confidence, UON confidence and disparity confidence. The disparity-based confidence is included due to its low computational cost and reliable free-space estimation with defined object boundaries [24]. Even though the FPHT algorithm takes the full disparity map as input, the individual confidence cues can be considered as independent, since the disparity map is used only as a coarse initialization for the FPHT hypothesis positions. Given this assumption, the resulting Stixel confidence $p(S)$ is computed as

$$p(S) = N \cdot p(S_{UON}) \cdot p(S_{FPHT}) \cdot p(D), \quad (1)$$

with $p(S_{UON})$ denoting the UON confidence, $p(S_{FPHT})$ the FPHT confidence, $p(D)$ the confidence derived from the disparity map, and $N$ the normalization term. The three confidence terms are properly normalized to obtain meaningful probabilities [24] with $p_{UOpr}$ being the prior probability to observe an obstacle Stixel:

$$N = p_{UOpr}/(p_{UOpr} \cdot p(S_{UON}) \cdot p(S_{FPHT}) \cdot p(D) + \\ (1 - p_{UOpr})(1 - p(S_{UON}))(1 - p(S_{FPHT}))(1 - p(D))). \quad (2)$$

We select $p_{UOpr} = 0.5$ deliberately to favor decisions towards obstacles, while the image statistics on the *Lost and Found* dataset suggest $p_{UOpr} < 0.01$.

**UON-Stixel Confidence:** The UON confidence is computed by the sum of the pixel-wise probability output of the non-free-space classes, i.e. $p(unexpected\ obst.) + p(background)$. We do not use the obstacle probability output directly, since e.g. pedestrians appearing in the dataset are often labeled as background and would be missed otherwise. The UON confidence is computed by

$$S_{UON} = \sum_{i \in S} (p_i(unexpected\ obst.) + p_i(background))/h_S, \quad (3)$$

denoting $h_S$ as the stixel height and $i$ as pixel index.

**FPHT-Stixel Confidence:** Similarly, for FPHT the available likelihood ratios are averaged for all point hypotheses contributing to the Stixel, denoted $l_{avg,FPHT}$, and then converted to a confidence via

$$p(S_{FPHT}) = 1/(1 + exp(l_{avg,FPHT})). \quad (4)$$

This yields similar probabilities for small Stixels with very few and large Stixels with many obstacle points. In order to incorporate prior information from the clustering step of FPHT, we also collect statistics of height and number of obstacle hypotheses for both true and false positive Stixels. Comparing the resulting probability density functions on the *Lost and Found* dataset, we observe an overlap below 50%, i.e. a reasonable separation, and obtain 10 obstacle points per Stixel and 10 cm height as the turning point when true positive Stixels obtain a higher probability than the false positives. This prior is multiplied onto $p(S_{FPHT})$, modeled by a sigmoidal function similar to [24].

**Disparity Confidence:** The disparity confidence estimation follows a hypothesis testing scheme. The energy (i.e. mean absolute disparity difference) for obstacle hypotheses (constant disparity) is compared against the free-space hypotheses (disparity slant according to the camera configura-

tion). These energies are estimated by

$$e_o = \sum_{i \in S} |d_i - \bar{d}|/h_S, \quad (5)$$

$$e_f = \sum_{i \in S} |d_i - \Delta d(r_{ctr} - r)|/h_S, \quad (6)$$

with $e_{o/f}$ denoting obstacle/free-space energy, $d_i$ the disparity within a Stixel $S$, $\bar{d}$ the mean disparity for the Stixel, $r_{ctr}$ the center row of the considered Stixel, and $r$ the row index of $d_i$. $\Delta d = B/H$ is the expected disparity slant of the road computed from baseline B and camera height over ground H. The disparity confidence is derived using the energy-probability relation [25] via

$$p(D) = 1/(1 + exp(e_o - e_f)). \quad (7)$$

The disparity variance is assumed to be 1.0 px. Patches of constant disparity are assigned high confidences whereas patches of the road surface obtain low confidences.

**Combining the Cues:** Final obstacle decisions are obtained by thresholding the fused Stixel confidences (Equation 1). If no UON-Stixel exists at the considered position, the UON confidence is computed according to Equation 3 since this information best reflects the available evidence for the existence of any obstacle type. If no FPHT Stixel is present we fall back to the neutral case $p(FPHT) = 0.5$. The complete fusion algorithm is summarized in Algorithm 1.

---

**Algorithm 1** Fusion of semantic and geometric cues

**Input**
  - UON heat maps, FPHT Stixels

**Output**
  - list of obstacle Stixels

**Algorithm**
 1: **function** COMPUTEUONSTIXELS( )
 2:   generate heat map in Stixel spacing
 3:   use UON result from argmax image as seed
 4:   travel along column until argmax label changes
 5:   **return** UON Stixels
 6: **end function**
 1: **function** FUSEUONFPHTSTIXELS( )
 2:   Find duplicate Stixels
 3:   OR-fuse unique Stixels
 4:   Apply UON confidence from heat map
 5:   Compute FPHT confidence
 6:   Compute disparity confidence
 7:   Perform Bayesian confidence estimation
 8:   **return** Fused Stixels with confidence
 9: **end function**

---

## IV. EVALUATION

### A. The Lost and Found Dataset and Extensions

For our evaluations we use the *Lost and Found* dataset introduced in [1]. It contains challenging scenarios including various small obstacle types, irregular road profiles, far object distances, different road surface appearance, and strong illumination changes. The data consists of 112 stereo video sequences at 2 MP resolution and pixel-accurate annotations for approximately 2100 frames.

**Distance and Height GT:** In addition to the available data, we generated object distance and height ground truth for all annotated frames. Initially, every stereo sequence set (per day and per location) is fine-calibrated for small relative roll, yaw, and pitch changes of the stereo camera.

Distance ground truth is obtained using the disparity data provided by Semi-Global Matching [23]. The disparity values within the labeled ground truth object borders are used to compute the mean and median disparity. The median is considered as ground truth if the mean deviates by less than 1 px, which is the case for more than 99.4% of the data.

Fig. 3 illustrates the distribution of the ground truth objects in the *Lost and Found* dataset regarding to their distances from the car, including instances of up to 200 m, with the majority of instances in the range of 10 to 70 m. The test set contains a maximum of 110 m.

With distance ground truth at hand, we interactively measured the height ground truth once per sequence for all objects under all recorded orientations using stereo data. The height-over-distance distribution (equivalent to object height in the image domain) is shown on the right side of Fig. 3. Since the focus lies on small objects, 97% of the instances have a height of less than 60 px.

### B. Application-Specific Evaluation Metrics

The performance evaluation presented in [1] covers both pixel- and instance-level. For the instance-level evaluation an instance-intersection value is computed yielding an overlap measure to ground truth. However, for automotive applications, object-level detection rates and corresponding false positive rates provide a more relevant performance measure.

Our object-level metric analyzes the overlap between the pixel-accurate annotations and the Stixel results of the proposed algorithms. A Stixel is defined as false positive if its overlap with the labeled free space area is larger than 50%, ignoring 10 px (0.25°) around ground truth obstacles, since errors caused by foreground fattening are considered acceptable for the considered task. Likewise, a true positive detection has more than 50% overlap with the ground truth label. An object is considered detected if one corresponding Stixel is found, admittedly an optimistic interpretation yielding an approximate upper bound for the detection rate.

### C. Quantitative Results

We evaluate detection rates and false positive rates of the algorithms (Section III) on the *Lost and Found* dataset. Fig. 4 depicts the detection rates over the distance range of the test dataset (5-110 m). The OR-fusion of semantic and geometric cues along with the probabilistic fusion ($p_{thresh} = 0.7$) exhibit the highest detection performance, yielding more than 90% for objects in ranges up to 50 m, followed by UON-Stixels, FPHT (with a better performance than UON for the

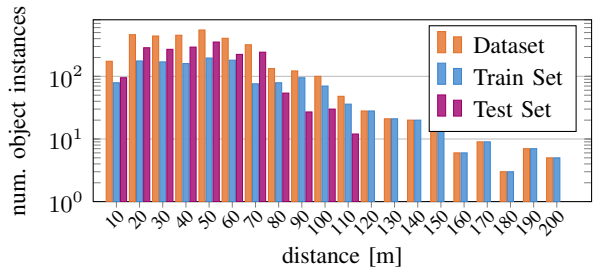

(a) Distribution of ground truth object instances over distance.

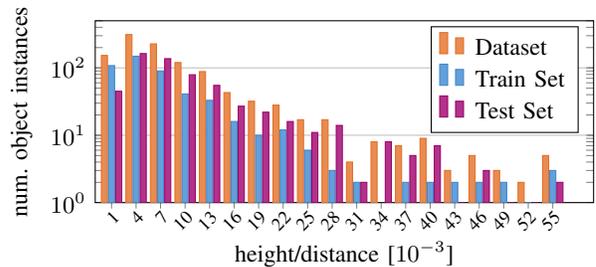

(b) Distribution of ground truth object instances over height/distance (object height in image). Height/distance=0.01 corresponds to 20 px.

Fig. 3. Statistics of the generated ground truth for the *Lost and Found* dataset [1].

|  | geom. | sem. | Detection Rate [%] | FP per frame | % frames with FP |
|---|---|---|---|---|---|
| Stixels [10] | ✓ | ✗ | 17.7 | 41.621 | 60.0 |
| Sem. Stixels [20] | ✓ | ✓ | 15.5 | 1.558 | 8.9 |
| FPHT-Stixels [1] | ✓ | ✗ | 55.4 | 0.573 | 28.0 |
| UON-Stixels | ✗ | ✓ | 73.8 | 0.103 | 5.3 |
| Fusion-AND | ✓ | ✓ | 42.1 | 0.009 | 0.3 |
| Fusion-OR | ✓ | ✓ | 84.1 | 0.669 | 30.4 |
| Fusion-Prob | ✓ | ✓ | 82.8 | 0.496 | 24.4 |

TABLE I

QUANTITATIVE RESULTS ON THE LOST AND FOUND DATASET.

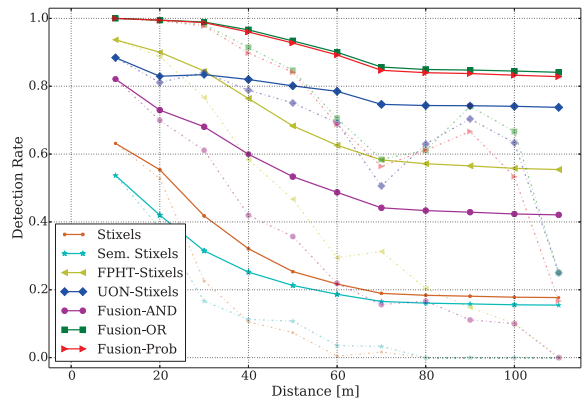

Fig. 4. Detection rate over distance. Solid curves correspond to integrated detection rate over distance while dashed curves represent the detection rate at each single distance range.

first 30 m), AND-fusion, ending with Stixels and Semantic Stixels. One can clearly see that the Stixel representation is not suitable for such small obstacles as present in the dataset. The UON-Stixels maintain a high detection rate throughout the distance range whereas FPHT-performance is best in the near to medium range where the objects cover a sufficient size in the image. For FPHT, a working point with 5 false positives per frame from [1] was chosen. This maps to 0.5 false positives per frame when the false positives within 10 px of the true objects are ignored (see Table I). UON-Stixels obtain 0.1 false positives per frame whereas the fusion result reaches rates comparable to FPHT alone but with a detection rate boost of about 50%. The Stixel representation yields the highest false positive rates mostly due to challenging 3D street profiles in the dataset. We conduct an additional evaluation considering height over distance, i.e. object height in the image (Fig. 5), where the FPHT algorithm performs best for objects beyond 20 px height where it obtains more than 90% detection rate. When comparing false positive Stixels with false positive Stixels frames, i.e. the number of frames containing at least one false positive Stixel (c.f. last two columns in Table I), one can see that the outrageous false positive rate of Stixels occur in a limited number of frames where the 3D road profile is not correctly estimated.

Fig. 6 shows false positive rates over detections rates for different confidence thresholds of the fusion approach. It can be seen that the detection rate is largely maintained for thresholds up to 0.7 with the confidence being an intuitive dial to reduce false positives. We also vary the obstacle prior ($p_{thresh} = 0.7$) and obtain a very similar curve (Fig. 6). Clearly, the best strategy is to keep all Stixel candidates and process them probabilistically in subsequent steps.

The false positive numbers listed here appear rather high, however, it is worth noting that the dataset was consciously designed to highlight rare challenges, including unusual drawings on the road surface and vertically curved road profiles. In practice, our fusion system delivers one false positive every 2 km driven. Moreover for FPHT-Stixels, false positives within the driving corridor appear about 4 times less frequent than false positives in the other image parts.

The whole system runs at 45 ms for SGM on FPGA, 35 ms for the CPU-optimized FPHT-Stixel computation (Intel Xeon 3GHz), 40 ms for the UON computation on the GPU (Nvidia TitanX), and 5 ms for the mid-level fusion on the CPU.

### D. Qualitative Results

Fig. 7 illustrates qualitative results of the proposed methods on three example sequences along with several baseline methods. The top row shows the respective input images, the second row displays the ground truth. The left column shows a scenario of lost cargo objects at around 40 m distance. Both Stixel baselines are unable to detect the object, FPHT detects the object but also a false positive due to a horizontal shadow and slight decalibration. UON cleanly detects all three objects and so does the fusion method without triggering on the false positive of FPHT.

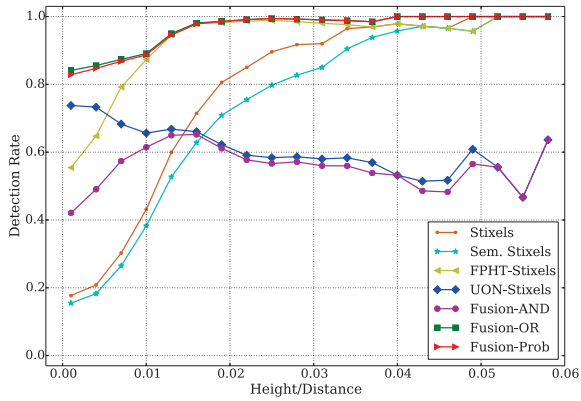

Fig. 5. Integrated detection rate over height/distance.

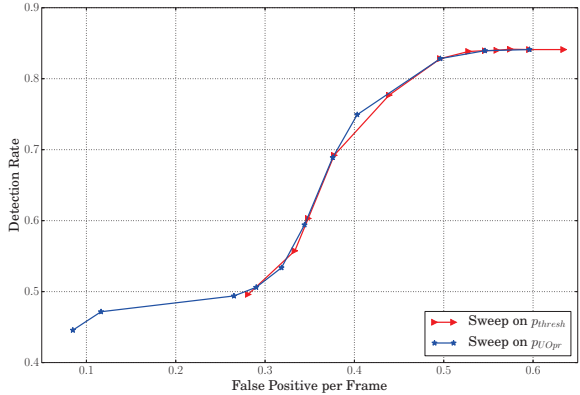

Fig. 6. Detection rate over false positive per frame for Fusion-Prob.

The middle column shows two small obstacles at 42 m distance with a height of only 5 cm, positioned on a road segment with strong vertical curvature. Here, the Stixel baseline generates many false positives due to inaccurate 3D road modeling, while Semantic Stixels correctly represent the road area but misses the objects. FPHT is unable to detect such small obstacles at this distance. UON and the fusion system both detect the obstacles without false positives.

In the right column, a scenario with a small child riding on a bobby car is depicted. Here, the Stixel baselines perform fairly well, FPHT cleanly detects the object but the UON method only yields a partial detection since both vehicles and pedestrians are part of the learned background class. The detection is correctly recovered in the fusion stage.

## V. CONCLUSIONS AND FUTURE WORK

In this work, we have presented a new framework for the detection of small unexpected obstacles for self-driving cars. We introduced the first deep learning system that successfully exploits visual appearance and contextual information to tackle this challenging task, achieving impressive results. Additionally, we have proposed a principled probabilistic fusion scheme with the currently best-performing stereo-based system, thus leveraging both semantic and geometric cues. We achieve a high-performance visual detection framework that proves to be robust to illumination changes, varying road appearance as well as 3D road profiles. Our system is able to detect critical obstacles of very low heights even at large distances. Compared to the state-of-the-art, our framework achieves a relative improvement of up to 50% in detection rate, while at the same time reducing false positives by 13%. The system operates in real-time on a self-driving platform.

As future work, we plan to collect additional training data to further increase the robustness of our system (e. g. recordings in bad weather) as well as directly include our probabilistic fusion scheme as part of our learning system.

**Acknowledgments:** We would like to thank Nicolai Schneider for his support during testing and evaluation.

## REFERENCES

[1] P. Pinggera, S. Ramos, S. Gehrig, U. Franke, C. Rother, and R. Mester, "Lost and Found: Detecting Small Road Hazards for Self-Driving Vehicles," in *IROS*, 2016.
[2] NHTSA, "Traffic safety facts 2011," NHTSA, Tech. Rep., 2011, URL:http://www-nrd.nhtsa.dot.gov/Pubs/811754AR.pdf.
[3] Y. LeCun, Y. Bengio, and G. Hinton, "Deep learning," *Nature*, 2015.
[4] Z. Zhang, R. Weiss, and A. Hanson, "Obstacle Detection Based on Qualitative and Quantitative 3D Reconstruction," *Trans. PAMI*, 1997.
[5] H. S. Sawhney, "3D Geometry from Planar Parallax," in *CVPR*, 1994.
[6] S. Kramm and A. Bensrhair, "Obstacle Detection Using Sparse Stereovision and Clustering Techniques," in *IV Symposium*, 2012.
[7] R. Labayrade, D. Aubert, and J.-P. Tarel, "Real Time Obstacle Detection in Stereovision on Non Flat Road Geometry Through "V-disparity" Representation," in *IV Symposium*, 2002.
[8] A. Wedel, H. Badino, C. Rabe, H. Loose, U. Franke, and D. Cremers, "B-Spline Modeling of Road Surfaces with an Application to Free Space Estimation," *Trans. ITS*, 2009.
[9] N. Bernini, M. Bertozzi, L. Castangia, M. Patander, and M. Sabbatelli, "Real-Time Obstacle Detection using Stereo Vision for Autonomous Ground Vehicles: A Survey," in *ITSC*, 2014.
[10] D. Pfeiffer and U. Franke, "Towards a global optimal multi-layer Stixel representation of dense 3D data," in *BMVC*, 2011.
[11] F. Oniga and S. Nedevschi, "Processing Dense Stereo Data Using Elevation Maps: Road Surface, Traffic Isle and Obstacle Detection," *Vehicular Technology*, 2010.
[12] R. Manduchi, A. Castano, A. Talukder, and L. Matthies, "Obstacle Detection and Terrain Classification for Autonomous Off-Road Navigation," *Autonomous Robots*, 2005.
[13] P. Fleischmann and K. Berns, *A Stereo Vision Based Obstacle Detection System for Agricultural Applications*. Springer, 2016, pp. 217–231.
[14] P. Pinggera, U. Franke, and R. Mester, "High-Performance Long Range Obstacle Detection Using Stereo Vision," in *IROS*, 2015.
[15] C. Creusot and A. Munawar, "Real-time Small Obstacle Detection on Highways using Compressive RBM Road Reconstruction," in *IV Symposium*, 2015.
[16] R. Hadsell, P. Sermanet, M. Scoffier, A. Erkan, K. Kavukcuoglu, U. Muller, and Y. LeCun, "Learning Long-Range Vision for Autonomous Off-Road Driving," *Journal of Field Robotics*, 2009.
[17] W. P. Sanberg, G. Dubbelman, and P. H. de With, "Free-Space Detection with Self-Supervised and Online Trained Fully Convolutional Networks," *arXiv:1604.02316v1 [cs.CV]*, 2016.
[18] J. Long, E. Shelhamer, and T. Darrell, "Fully Convolutional Networks for Semantic Segmentation," in *CVPR*, 2015.
[19] A. Bendale and T. E. Boult, "Towards Open Set Deep Networks," in *CVPR*, 2016.
[20] L. Schneider, M. Cordts, T. Rehfeld, D. Pfeiffer, M. Enzweiler, U. Franke, M. Pollefeys, and S. Roth, "Semantic Stixels: Depth is Not Enough," in *IV Symposium*, 2016.
[21] C. Szegedy, W. Liu, Y. Jia, P. Sermanet, S. Reed, D. Anguelov, D. Erhan, V. Vanhoucke, and A. Rabinovich, "Going Deeper with Convolutions," in *CVPR*, 2015.
[22] M. Cordts, M. Omran, S. Ramos, T. Rehfeld, M. Enzweiler, R. Benenson, U. Franke, S. Roth, and B. Schiele, "The Cityscapes Dataset for Semantic Urban Scene Understanding," in *CVPR*, 2016.
[23] S. Gehrig, R. Stalder, and N. Schneider, "A Flexible High-Resolution Real-Time Low-Power Stereo Vision Engine," in *ICVS*, 2015.
[24] S. Gehrig, A. Barth, N. Schneider, and J. Siegemund, "A multi-cue approach for stereo-based object confidence estimation," in *IROS*, 2012.
[25] R. Gray, *Entropy and information theory*. Springer, 1990.

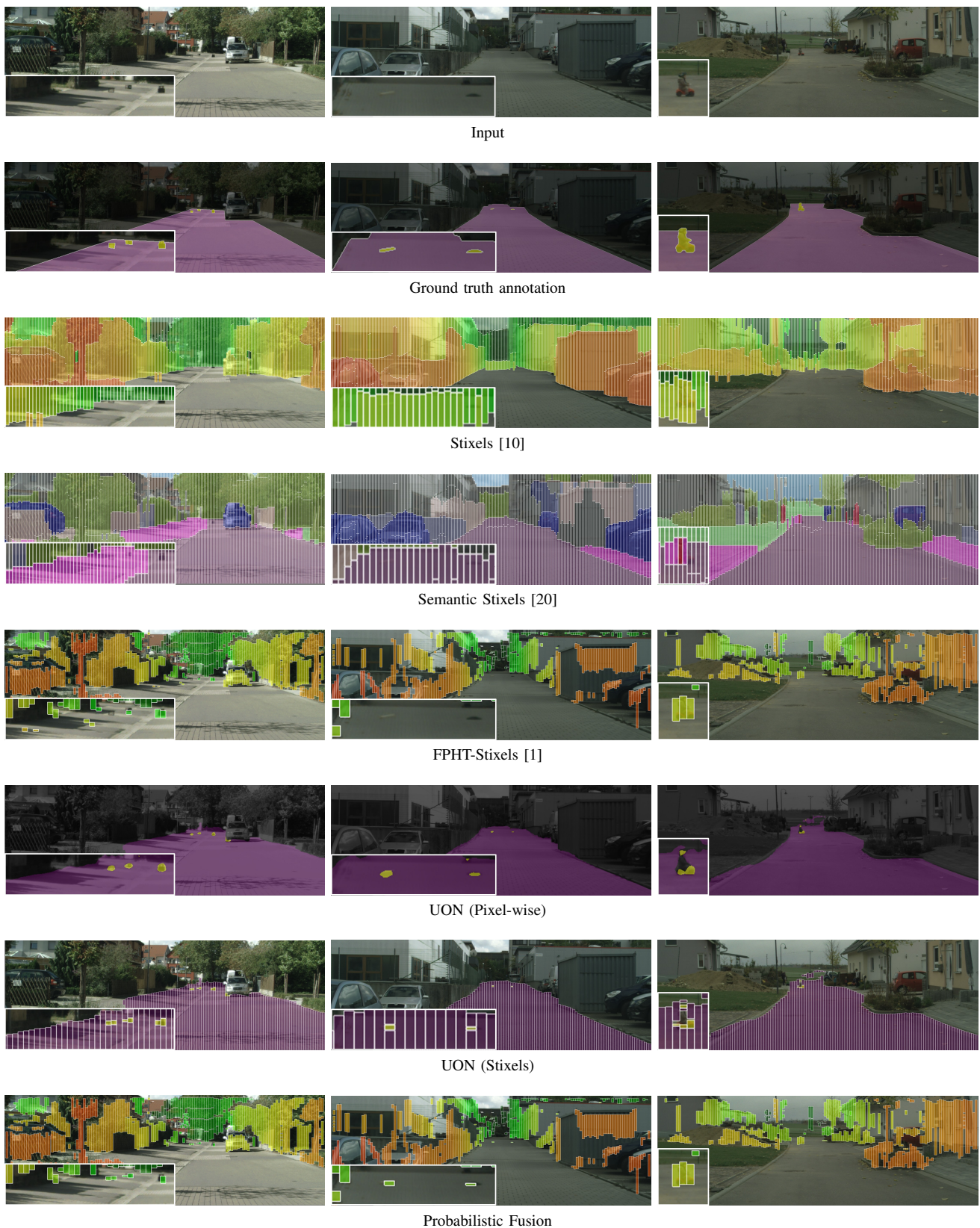

Fig. 7. Qualitative results of the evaluated methods. The top row shows the left input image, lower rows show pixel-wise and mid-level detections as overlay, color-coded by distance (red: near, green: far). For the semantic output, the color purple denotes free-space while the color gold denotes obstacles.